\pdfoutput=1

\documentclass[11pt]{article}

\usepackage{acl}

\usepackage{times}
\usepackage{latexsym}

\usepackage[utf8]{inputenc}

\usepackage{url}
\usepackage[caption=false]{subfig}
\usepackage{graphicx}
\usepackage{kotex}

\usepackage{hyperref}
\definecolor{darkblue}{rgb}{0, 0, 0.5}
\hypersetup{colorlinks=true,citecolor=darkblue, linkcolor=darkblue, urlcolor=darkblue}

\usepackage{color}
\usepackage{url}

\usepackage[linguistics]{forest} 
\usepackage{synttree}
\usepackage{tikz-dependency}
\usepackage{colortbl}

\title{Yet Another Format of Universal Dependencies for Korean}

\author{
Yige Chen$^{\dagger}$\thanks{~Yige Chen, Eunkyul Leah Jo, and Yundong Yao contributed equally.} ~~~~ 
Eunkyul Leah Jo$^{\ddagger*}$ ~~~ 
Yundong Yao$^{\ddagger*}$ \\
\textbf{KyungTae Lim$^{\mathparagraph}$ ~~ 
Miikka Silfverberg$^{\ddagger}$ ~~ 
Francis M. Tyers$^{\mathsection}$ ~~ 
Jungyeul Park$^{\ddagger}$}\\
$^{\dagger}$The Chinese University of Hong Kong, Hong Kong ~~ $^{\ddagger}$The University of British Columbia, Canada\\
$^{\mathparagraph}$Hanbat National University \& TeddySum, South Korea ~~ $^{\mathsection}$Indiana University, USA\\
$^{\dagger}${\tt{yigechen@link.cuhk.edu.hk}}~ $^{\ddagger}${\tt\{eunkyul,iameleve\}@student.ubc.ca}\\ 
{\tt $^{\mathparagraph}$ktlim@hanbat.ac.kr} ~ $^{\mathsection}${\tt{ftyers@iu.edu}} ~ $^{\ddagger}$\{{\tt{msilfver,jungyeul}\}@mail.ubc.ca} \\
}

\begin{document}
\maketitle
\begin{abstract}
In this study, we propose a morpheme-based scheme for Korean dependency parsing and adopt the proposed scheme to Universal Dependencies. We present the linguistic rationale that illustrates the motivation and the necessity of adopting the morpheme-based format, and develop scripts that convert between the original format used by Universal Dependencies and the proposed morpheme-based format automatically. The effectiveness of the proposed format for Korean dependency parsing is then testified by both statistical and neural models, including UDPipe and  Stanza, with our carefully constructed morpheme-based word embedding for Korean. \textsc{morphUD} outperforms parsing results for all Korean UD treebanks, and we also present detailed error analyses. 

\end{abstract}

\section{Introduction} \label{intro}
Dependency parsing is one of the tasks in natural language processing that have been investigated extensively. Using the dependency grammar, it finds the relations between the words in a sentence, and forms an acyclic dependency graph that explains the grammatical structure of the sentence.
With various machine learning techniques, the current dependency parsers are able to approach human performances given English corpora.\footnote{\url{https://ai.googleblog.com/2016/05/announcing-syntaxnet-worlds-most.html}} Previous studies trying to probe or improve Korean dependency parsers are lacking, especially for those considering the underlying reason and rationale based on the linguistic properties of Korean. The parsers do not perform as well as their English counterparts, partly due to the fact that Korean is a language that has more complicated linguistic features that make parsing on the word level difficult. 

There have been previous studies trying to cope with the word-level representation issues of Korean \citep{choi-palmer:2011:SPMRL,park-EtAl:2013:IWPT,kanayama-EtAl:2014:LT4CloseLang}. Given that Korean is an agglutinative language that heavily relies on morphemes, and the natural segmentation does not correctly reflect either the words or the morphemes of Korean texts, \citet{park-tyers:2019:LAW} suggested an annotation scheme that decomposes Korean texts into the morpheme level, and applied the morpheme-based format to POS tagging. 

In this study, we propose a morpheme-based scheme for Korean dependency parsing that is developed based on \citet{park-tyers:2019:LAW}, and adopt the proposed scheme to Universal Dependencies \citep{nivre-EtAl:2016:LREC,nivre-EtAl:2020:LREC}, which contains two Korean dependency parsing treebanks, namely the GSD treebank \citep{mcdonald-EtAl:2013:ACL} and the Kaist treebank \citep{choi-EtAl:1994,chun-EtAl:2018:LREC}. While the two Korean treebanks meet the standards of Universal Dependencies and have been studied for dependency parsing tasks extensively \citep{kondratyuk-straka-2019-75,qi-EtAl:2020:ACL}, the treebanks are formatted in a way that the natural segmentations of Korean texts are preserved, and even with some morpheme-level information, only the language-specific part-of-speech tags on the morpheme level are included { in the treebanks}, and {both treebanks do} not have any morpheme-level parsing tags. Different from the traditional scheme based on natural segmentation, this scheme utilizes the inherent morphological and typological features of the Korean language, and the morpheme-level parsing tags can therefore be derived using a set of linguistically motivated rules, which are further used to produce the morpheme-level dependency parsing results and automatic conversions between the morpheme-based format and the traditional format.

The proposed morpheme-based representation is examined using several dependency parsing models, including UDPipe \citep{straka-hajic-strakova:2016:LREC,straka-strakova:2017:CoNLL} and Stanza \citep{qi-EtAl:2020:ACL}. Compared to the baseline models trained using the two treebanks without modification, our proposed format makes statistically significant improvements in the performances of the parsing models for the Korean language as reported in the error analysis. 


\section{Representation of {\textsc{morphUD}}} \label{section-represent-morphud-korean}

\begin{figure*}
\centering
{\tiny
\begin{dependency}
\begin{deptext}
프랑스 \&의 \& 세계 \&적 \&이 \&ㄴ  \& 의상  \&디자이너 \&엠마누엘  \&웅가로 \&가 \& 실내 \& 장식 \&용 \& 직물  \& 디자이너 \&로  \& 나서 \&었 \&다  \&. \\
 \textit{peurangseu} \& \textit{-ui} \&  \textit{segye} \& \textit{-jeok} \& \textit{-i} \& \textit{-n} \&
 \textit{uisang} \& \textit{dijaineo} \& \textit{emmanuel} \& \textit{unggaro} \& \textit{-ga} \& 
 \textit{silnae} \&  \textit{jangsik} \& \textit{-yong} \& 
 \textit{jikmul} \& \textit{dijaineo} \& \textit{-ro} \& 
 \textit{naseo} \& \textit{-eoss} \& \textit{-da} \& \textit{.}\\ 
 France \& -\textsc{gen} \& world \& -\textsc{suf} \& -\textsc{cop}\& -\textsc{rel} \& 
 fashion \& designer \& Emanuel \& Ungaro \& -\textsc{nom} \& 
 interior \& decoration \& usage \& textile \& designer \& -\textsc{ajt} \& 
 become \& -\textsc{past} \& -\textsc{ind} \& {.}\\
1 \& 2 \& 3 \& 4 \& 5 \& 6 \& 7 \& 8 \& 9 \& 10 \& 11 \& 12 \& 13 \& 14 \& 15 \& 16 \& 17 \& 18 \& 19 \& 20 \& 21\\
\end{deptext}
   \depedge[edge unit distance=1.7ex]{1}{8}{nmod} 
   \depedge[edge below]{2}{1}{case}
   \depedge[edge unit distance=1.5ex]{3}{8}{acl} 
   \depedge[edge below]{4}{3}{aux}
   \depedge[edge below]{5}{3}{aux}
   \depedge[edge below]{6}{3}{aux}
   \depedge{7}{8}{compound}
   \depedge{8}{10}{compound}
   \depedge{9}{10}{compound}
   \depedge[edge unit distance=0.96ex]{10}{18}{nsubj} 
   \depedge[edge below]{11}{10}{case}
   \depedge{12}{13}{compound} 
   \depedge[edge unit distance=2.6ex]{13}{15}{compound} 
   \depedge[edge below]{14}{13}{aux}
   \depedge{15}{16}{compound} 
   \depedge[edge unit distance=1.9ex]{16}{18}{advcl} 
   \depedge[edge below]{17}{16}{case}
   \deproot{18}{root} 
   \depedge[edge below]{19}{18}{aux}
   \depedge[edge below]{20}{18}{aux}
   \depedge[edge unit distance=1.3ex]{21}{18}{punct} 
\end{dependency}
}
\caption{Example of morpheme-based universal dependencies for Korean: while dependencies in top-side are the original dependencies between words, dependencies in bottom-side are newly added dependencies for between morphemes.}
\label{morphUD-example}
\end{figure*}

In this study, we adopt a morpheme-based format that captures the linguistic properties of the Korean language proposed by \citet{park-tyers:2019:LAW}. The natural segmentation of Korean is based on eojeol, which does not necessarily reflect the actual word or morpheme boundaries of the language. For example, an eojeol of Korean may contain both a noun and its postposition, or both a verb and its particles marking tense, aspect, honorifics, etc. While this is typical for Korean as an agglutinative language, it creates difficulties and challenges for NLP tasks regarding the Korean language, including dependency parsing. It is not ideal that the tokens dependency relations are annotated on are sometimes words, and sometimes phrases as an eojeol may consist of more than a word. Furthermore, Korean as an agglutinative language has very regular conjugations, which makes it easy and natural to split those words and phrases into morphemes when analyzing the language since nearly every piece of an eojeol can be identified to be of a certain meaning or function.

The morpheme-based format aims at decomposing the Korean sentences further into morphemes, which means that dependency relations are no longer marked on the eojeol level. Instead, they are marked on morphemes such that within each eojeol that is not monomorphemic, a head of that eojeol will be found and all other morphemes will be attached directly to the head. As a result, the head of a non-monomorphemic eojeol carries the dependency relation this eojeol originally has, and all other morphemes will be attached to it. In order to find the head, we develop a script and apply some heuristics which include that the head of an eojeol is usually a noun, a proper noun, or a verb, and while there is no noun or verb in an eojeol, the script we implemented continues to find other morphemes such as pronouns, adjectives, adverbs, numerals, etc. The script also excludes the use of adpositions, conjunctions, and particles as heads in most cases, unless these are the only part-of-speeches in an eojeol except for punctuations. While there are multiple morphemes that can be heads in an eojeol, the script will decide which one to take based on the part-of-speeches of the morphemes. For instance, when there are multiple nouns, the last one will carry the dependency relation of the eojeol, whereas when there are multiple verbs, the first verb will carry the dependency relation as Korean is a head-final language. Once the head of a non-monomorphemic eojeol is found, the other morphemes will be dependent on the head and be assigned with other dependency relations such as compound, case, auxiliary depending on their UPOS and XPOS.

\section{Experiments and results} \label{section-experiments-results}

\subsection{Data and systems}
In this study, we deploy two parsers to evaluate our proposed format, namely UDPipe \citep{straka-hajic-strakova:2016:LREC} as a baseline system and Stanza \citep{qi-EtAl:2020:ACL} as one of the {state-of-the-art} dependency parsers. 
UDPipe is a pipeline designed for processing CoNLL-U formatted files, which performs tokenization using Bi-LSTM, morphological analysis, part-of-speech tagging, lemmatization using MorphoDiTa \citep{strakova-etal-2014-open}, and dependency parsing using slightly modified Parsito \citep{straka-parsing-2015}. Since the whole pipeline needs no language-specific knowledge, which means that it can be trained using corpora in a different scheme, we choose UDPipe as our baseline. 
Stanza is another natural language processing toolkit that includes Dozat's biaffine attention dependency parser \citep{dozat-manning:2017:ICLR}. Dozat's dependency parser uses the minimum spanning tree algorithm that can deal with non-projectivity dependency relations, and more importantly it excelled all of dependency parsers during CoNLL 2017 and 2018 Shared Task \citep{zeman-EtAl:2017:CoNLL,zeman-EtAl:2018:CoNLL}
In this study, the two dependency parsing pipelines take both the original word-based form\footnote{While words and eojeols are not the same in Korean based on their definitions, in this study, the terms ``word-based'' and ``eojeol-based'' are interchangeable. } (the current scheme adopted by Universal Dependencies), which we denote as \textsc{wordUD}, and the morpheme-based form, which we denote as \textsc{morphUD}, of the GSD and KAIST treebanks as the input.

We develop the script to convert between the \textsc{wordUD} format and our proposed \textsc{morphUD} format. The script consists of two major components, which are \textsc{wordUD} to \textsc{morphUD} (\texttt{Word2Morph}) and \textsc{morphUD} to \textsc{wordUD} (\texttt{Morph2Word}). The Word2Morph component splits the word tokens in the CoNLL-U treebank of Korean into morphemes using the lemmas already provided, and assigns dependency relations on the resegmented tokens based on the original dependency relations annotated on the word tokens. 
The Morph2Word component, on the other hand, firstly pairs the tokens in the \textsc{wordUD} dataset and the \textsc{morphUD} dataset, and then assigns the dependency relations from morpheme tokens in \textsc{morphUD} to word tokens in \textsc{wordUD}. 
Within both components, a root detector for the word is implemented in order to find the root (or stem) of a word when the word is multimorphemic (i.e., needs to be split into morphemes and attach dependency relations on it correspondingly). { Evaluations of the conversion scripts are not conducted in this study, since the morphemes are inherited from the lemmas in the treebanks, and the part-of-speech tags, roots, and dependency relations are predicted and assigned to the morphemes based on the linguistic features and the grammar of Korean that are regular, as presented in Section \ref{section-represent-morphud-korean}.}

\subsection{Results}
We report the labeled attachment score (LAS), which is a standard evaluation metric in dependency parsing, using the evaluation script (2018 version) provided by CoNLL 2018 Shared Task.\footnote{\url{https://universaldependencies.org/conll18/conll18_ud_eval.py}}
Table~\ref{results-main} shows results of udpipe as a baseline system and stanza as one of the state-of-the-art systems. 
All results are reported in the \textsc{wordUD} format. 
That is, all experiments are trained and predicted in the proposed \textsc{morphUD} format, and then the result is converted back to the \textsc{wordUD} format for comparison purposes. 
We train udpipe once because it can produce the same parsing model if we train it on the same machine.
For stanza, we provide average LAS and its standard deviation after five training and evaluation. 
Both systems use the finely crafted 300d embedding file by fastText \citep{bojanowski-EtAl:2017:TACL}: \textsc{wordUD} and \textsc{morphUD} use words and morphemes as their embedding entries, respectively to make sure that their input representation would be correctly matched.
For embeddings, there are 9.6M sentences and 157M words (tokenized)  based on \textsc{wordUD}. The set of documents for embeddings includes all articles published in \textit{The Hankyoreh} during 2016 (1.2M sentences), Sejong morphologically analyzed corpus (3M), and Korean Wikipedia articles (20201101) (5.3M).
As expected, all results of \textsc{morphUD} outperform \textsc{wordUD} in Table~\ref{results-main}.

\begin{table*}
\centering
{
\begin{tabular}{|r | cc | cc|} \hline
& \multicolumn{2}{c|}{\texttt{ko\_gsd}} &  \multicolumn{2}{c|}{\texttt{ko\_kaist}} \\
& \textsc{wordUD} 
& \textsc{+morphUD}
& \textsc{wordUD} 
& \textsc{+morphUD}\\ \hline
udpipe    & 70.90 
& 77.01 
& 77.01  
& 81.80 \\
stanza   & 84.63 ($\pm0.18$) 
& 84.98 ($\pm0.20$) 
& 86.67 ($\pm0.17$) 
& 88.46 ($\pm0.14$)\\ \hline
\end{tabular}
}
\caption{Dependency parsing results: for the comparison purpose all \textsc{morphUD} results are converted back to \textsc{wordUD} after training and predicting with the format of \textsc{morphUD}}
\label{results-main}
\end{table*}

\subsection{Error analysis and discussion}

Figure~\ref{confusion-matrix-direction} shows the confusion matrix between \textsc{wordUD} and \textsc{morphUD}, in which the column and the row represent the arc direction of \texttt{gold} and \texttt{system}, respectively. 
\textsc{morphUD} outperforms \textsc{wordUD} in predicting all directions except for right (\texttt{gold}) / left (\texttt{system}).
The system predicts the left arc instead of the correct right arc (212 arc direction errors in \textsc{wordUD} vs. 240 in \textsc{morphUD}). This is because we spuriously added left arcs for functional morphemes in \textsc{morphUD} where the system learned more left arc instances during training. 
\begin{figure}[t]
\centering
\subfloat[b][\textsc{wordUD}]{
{
\begin{tabular}{c ccc}
& \textsc{l}& \textsc{r} & \textsc{o} \\
\textsc{l} & 
\cellcolor{blue!0} & 
\cellcolor{blue!100.0} & 
\cellcolor{blue!28.482972136222912}\\
\textsc{r} &  
\cellcolor{blue!65.63467492260062} & 
\cellcolor{blue!0} & 
\cellcolor{blue!54.79876160990712} \\
\textsc{o} &  
\cellcolor{blue!57.27554179566563} & 
\cellcolor{blue!26.006191950464398} & 
\cellcolor{blue!0}\\
\end{tabular}
}\label{confusion-wordud-dir}}
\subfloat[b][\textsc{morphUD}]{
{
\begin{tabular}{c ccc}
& \textsc{l}& \textsc{r} & \textsc{o} \\
\textsc{l} & 
\cellcolor{blue!0} & 
\cellcolor{blue!73.37461300309597} & 
\cellcolor{blue!27.55417956656347} \\
\textsc{r} &  
\cellcolor{blue!74.30340557275542} & 
\cellcolor{blue!0} & 
\cellcolor{blue!52.94117647058824} \\

\textsc{o} &  
\cellcolor{blue!54.79876160990712} & 
\cellcolor{blue!25.696594427244584} & 
\cellcolor{blue!0} \\
\end{tabular}
}\label{confusion-morphud-dir}}

\caption{Confusion matrix for the direction of arcs where the column represents \texttt{gold}, and the row \texttt{system}: \textsc{l}eft, \textsc{r}ight, and \textsc{o} for \textsc{to root}.}
\label{confusion-matrix-direction}
\end{figure}

\begin{figure}[t]
\centering
\subfloat[b][\textsc{wordUD}]{
\resizebox{0.23\textwidth}{!}{
\begin{tabular}{c ccccc ccccc c}
  & 0 & 1 & 2 & 3 & 4 & 5 & 6 & 7 & 8 & 9 & 10 \\
0 & 
\cellcolor{red!0} & \cellcolor{red!23.52941176470588} &
\cellcolor{red!5.216426193118757} & \cellcolor{red!0.6659267480577136} &
\cellcolor{red!0.4439511653718091} &  \cellcolor{red!0} & 
\cellcolor{red!0} & \cellcolor{red!0} & 
\cellcolor{red!0} & \cellcolor{red!0} & \cellcolor{red!0} \\

1 & \cellcolor{red!24.30632630410655} & \cellcolor{red!0} & 
\cellcolor{red!77.80244173140954} & \cellcolor{red!19.644839067702552} & 
\cellcolor{red!5.105438401775805} & \cellcolor{red!2.2197558268590454} & 
\cellcolor{red!0.9988901220865706} & \cellcolor{red!0.3329633740288568} & 
\cellcolor{red!0} & \cellcolor{red!0} & \cellcolor{red!0} \\

2 & \cellcolor{red!4.439511653718091} &
\cellcolor{red!100.0} &
\cellcolor{red!0.0} &
\cellcolor{red!65.92674805771365} &
\cellcolor{red!19.20088790233074} &
\cellcolor{red!5.216426193118757} &
\cellcolor{red!2.774694783573807} &
\cellcolor{red!0.22197558268590456} &
\cellcolor{red!0.22197558268590456} &
\cellcolor{red!0.0} \\

3 & \cellcolor{red!0.4439511653718091} &
\cellcolor{red!25.527192008879023} &
\cellcolor{red!96.11542730299666} &
\cellcolor{red!0.0} &
\cellcolor{red!45.39400665926748} &
\cellcolor{red!16.75915649278579} &
\cellcolor{red!3.662597114317425} &
\cellcolor{red!2.108768035516093} &
\cellcolor{red!0.11098779134295228} &
\cellcolor{red!0.22197558268590456} &
\cellcolor{red!0.0} \\

4 & \cellcolor{red!0.3329633740288568} &
\cellcolor{red!8.102108768035517} &
\cellcolor{red!23.307436182019977} &
\cellcolor{red!69.47835738068812} &
\cellcolor{red!0.0} &
\cellcolor{red!26.304106548279687} &
\cellcolor{red!10.321864594894562} &
\cellcolor{red!2.2197558268590454} &
\cellcolor{red!2.2197558268590454} &
\cellcolor{red!0.0} &
\cellcolor{red!0.22197558268590456} \\

5 & \cellcolor{red!0.22197558268590456} &
\cellcolor{red!2.885682574916759} &
\cellcolor{red!6.8812430632630415} &
\cellcolor{red!16.537180910099888} &
\cellcolor{red!34.517203107658155} &
\cellcolor{red!0.0} &
\cellcolor{red!13.540510543840178} &
\cellcolor{red!5.105438401775805} &
\cellcolor{red!2.108768035516093} &
\cellcolor{red!1.2208657047724751} &
\cellcolor{red!0.0} \\

6 & \cellcolor{red!0.11098779134295228} &
\cellcolor{red!0.776914539400666} &
\cellcolor{red!1.6648168701442843} &
\cellcolor{red!4.328523862375139} &
\cellcolor{red!9.100998890122087} &
\cellcolor{red!13.540510543840178} &
\cellcolor{red!0.0} &
\cellcolor{red!4.661487236403995} &
\cellcolor{red!0.8879023307436182} &
\cellcolor{red!0.9988901220865706} &
\cellcolor{red!0.0} \\

7 & \cellcolor{red!0.0} &
\cellcolor{red!0.3329633740288568} &
\cellcolor{red!0.3329633740288568} &
\cellcolor{red!0.9988901220865706} &
\cellcolor{red!1.4428412874583796} &
\cellcolor{red!3.107658157602664} &
\cellcolor{red!5.660377358490567} &
\cellcolor{red!0.0} &
\cellcolor{red!0.9988901220865706} &
\cellcolor{red!0.0} &
\cellcolor{red!0.3329633740288568} \\

8 & \cellcolor{red!0.0} &
\cellcolor{red!0.0} &
\cellcolor{red!0.0} &
\cellcolor{red!0.0} &
\cellcolor{red!0.6659267480577136} &
\cellcolor{red!0.3329633740288568} &
\cellcolor{red!0.5549389567147613} &
\cellcolor{red!1.8867924528301887} &
\cellcolor{red!0.0} &
\cellcolor{red!0.0} &
\cellcolor{red!0.0} \\

9 & \cellcolor{red!0.0} &
\cellcolor{red!0.0} &
\cellcolor{red!0.0} &
\cellcolor{red!0.11098779134295228} &
\cellcolor{red!0.0} &
\cellcolor{red!0.0} &
\cellcolor{red!0.0} &
\cellcolor{red!0.11098779134295228} &
\cellcolor{red!0.11098779134295228} &
\cellcolor{red!0.0} &
\cellcolor{red!0.0} \\

10 & \cellcolor{red!0.0} &
\cellcolor{red!0.0} &
\cellcolor{red!0.0} &
\cellcolor{red!0.0} &
\cellcolor{red!0.22197558268590456} &
\cellcolor{red!0.0} &
\cellcolor{red!0.0} &
\cellcolor{red!0.0} &
\cellcolor{red!0.0} &
\cellcolor{red!0.0} &
\cellcolor{red!0.0} \\
\end{tabular}
}
\label{confusion-wordud}}
\subfloat[b][\textsc{morphUD}]{
\resizebox{0.23\textwidth}{!}{
\begin{tabular}{c ccccc ccccc c}
  & 0 & 1 & 2 & 3 & 4 & 5 & 6 & 7 & 8 & 9 & 10 \\
0 & 
\cellcolor{red!0.0} &
\cellcolor{red!22.97447280799112} &
\cellcolor{red!4.550499445061043} &
\cellcolor{red!1.2208657047724751} &
\cellcolor{red!0.11098779134295228} &
\cellcolor{red!0.0} &
\cellcolor{red!0.0} &
\cellcolor{red!0.0} &
\cellcolor{red!0.0} &
\cellcolor{red!0.0} &
\cellcolor{red!0.0} \\

1 &
\cellcolor{red!23.52941176470588} &
\cellcolor{red!0.0} &
\cellcolor{red!80.6881243063263} &
\cellcolor{red!18.756936736958934} &
\cellcolor{red!5.660377358490567} &
\cellcolor{red!1.2208657047724751} &
\cellcolor{red!0.5549389567147613} &
\cellcolor{red!0.3329633740288568} &
\cellcolor{red!0.0} &
\cellcolor{red!0.0} &
\cellcolor{red!0.0} \\

2 & 
\cellcolor{red!3.9955604883462823} &
\cellcolor{red!94.89456159822419} &
\cellcolor{red!0.0} &
\cellcolor{red!66.48168701442842} &
\cellcolor{red!18.201997780244174} &
\cellcolor{red!5.549389567147614} &
\cellcolor{red!0.8879023307436182} &
\cellcolor{red!0.22197558268590456} &
\cellcolor{red!0.22197558268590456} &
\cellcolor{red!0.0} &
\cellcolor{red!0.0} \\

3 & 
\cellcolor{red!0.8879023307436182} &
\cellcolor{red!24.195338512763595} &
\cellcolor{red!91.00998890122086} &
\cellcolor{red!0.0} &
\cellcolor{red!45.94894561598224} &
\cellcolor{red!15.760266370699222} &
\cellcolor{red!3.4406215316315207} &
\cellcolor{red!1.2208657047724751} &
\cellcolor{red!0.3329633740288568} &
\cellcolor{red!0.11098779134295228} &
\cellcolor{red!0.0} \\

4 & 
\cellcolor{red!0.22197558268590456} &
\cellcolor{red!8.76803551609323} &
\cellcolor{red!27.081021087680355} &
\cellcolor{red!64.37291897891232} &
\cellcolor{red!0.0} &
\cellcolor{red!23.52941176470588} &
\cellcolor{red!10.099889012208656} &
\cellcolor{red!2.5527192008879025} &
\cellcolor{red!1.3318534961154271} &
\cellcolor{red!0.0} &
\cellcolor{red!0.0} \\

5 & 
\cellcolor{red!0.22197558268590456} &
\cellcolor{red!2.108768035516093} &
\cellcolor{red!6.992230854605993} &
\cellcolor{red!19.08990011098779} &
\cellcolor{red!31.853496115427305} &
\cellcolor{red!0.0} &
\cellcolor{red!12.430632630410656} &
\cellcolor{red!3.884572697003329} &
\cellcolor{red!1.7758046614872365} &
\cellcolor{red!0.6659267480577136} &
\cellcolor{red!0.0} \\

6 & 
\cellcolor{red!0.0} &
\cellcolor{red!0.9988901220865706} &
\cellcolor{red!1.9977802441731412} &
\cellcolor{red!3.3296337402885685} &
\cellcolor{red!9.100998890122087} &
\cellcolor{red!13.762486126526083} &
\cellcolor{red!0.0} &
\cellcolor{red!5.105438401775805} &
\cellcolor{red!0.8879023307436182} &
\cellcolor{red!0.5549389567147613} &
\cellcolor{red!0.0}  \\

7 & 
\cellcolor{red!0.0} &
\cellcolor{red!0.0} &
\cellcolor{red!0.6659267480577136} &
\cellcolor{red!0.8879023307436182} &
\cellcolor{red!1.9977802441731412} &
\cellcolor{red!3.7735849056603774} &
\cellcolor{red!4.772475027746948} &
\cellcolor{red!0.0} &
\cellcolor{red!0.8879023307436182} &
\cellcolor{red!0.11098779134295228} &
\cellcolor{red!0.11098779134295228} \\

8 & 
\cellcolor{red!0.0} &
\cellcolor{red!0.0} &
\cellcolor{red!0.0} &
\cellcolor{red!0.22197558268590456} &
\cellcolor{red!0.6659267480577136} &
\cellcolor{red!0.5549389567147613} &
\cellcolor{red!1.4428412874583796} &
\cellcolor{red!1.2208657047724751} &
\cellcolor{red!0.0} &
\cellcolor{red!0.0} &
\cellcolor{red!0.0}  \\

9 & 
\cellcolor{red!0.0} &
\cellcolor{red!0.0} &
\cellcolor{red!0.0} &
\cellcolor{red!0.11098779134295228} &
\cellcolor{red!0.0} &
\cellcolor{red!0.11098779134295228} &
\cellcolor{red!0.0} &
\cellcolor{red!0.0} &
\cellcolor{red!0.4439511653718091} &
\cellcolor{red!0.0} &
\cellcolor{red!0.0} \\

10 & 
\cellcolor{red!0.0} &
\cellcolor{red!0.0} &
\cellcolor{red!0.0} &
\cellcolor{red!0.0} &
\cellcolor{red!0.22197558268590456} &
\cellcolor{red!0.0} &
\cellcolor{red!0.0} &
\cellcolor{red!0.0} &
\cellcolor{red!0.0} &
\cellcolor{red!0.0} &
\cellcolor{red!0.0} \\
\end{tabular}
}\label{confusion-morphud}}

\caption{Confusion matrix for the depth of arcs where the column represents \texttt{gold}, and the row \texttt{system}.}
\label{confusion-matrix}
\end{figure}

\begin{figure}[t]
\centering
\resizebox{.4\textwidth}{!}{
\includegraphics{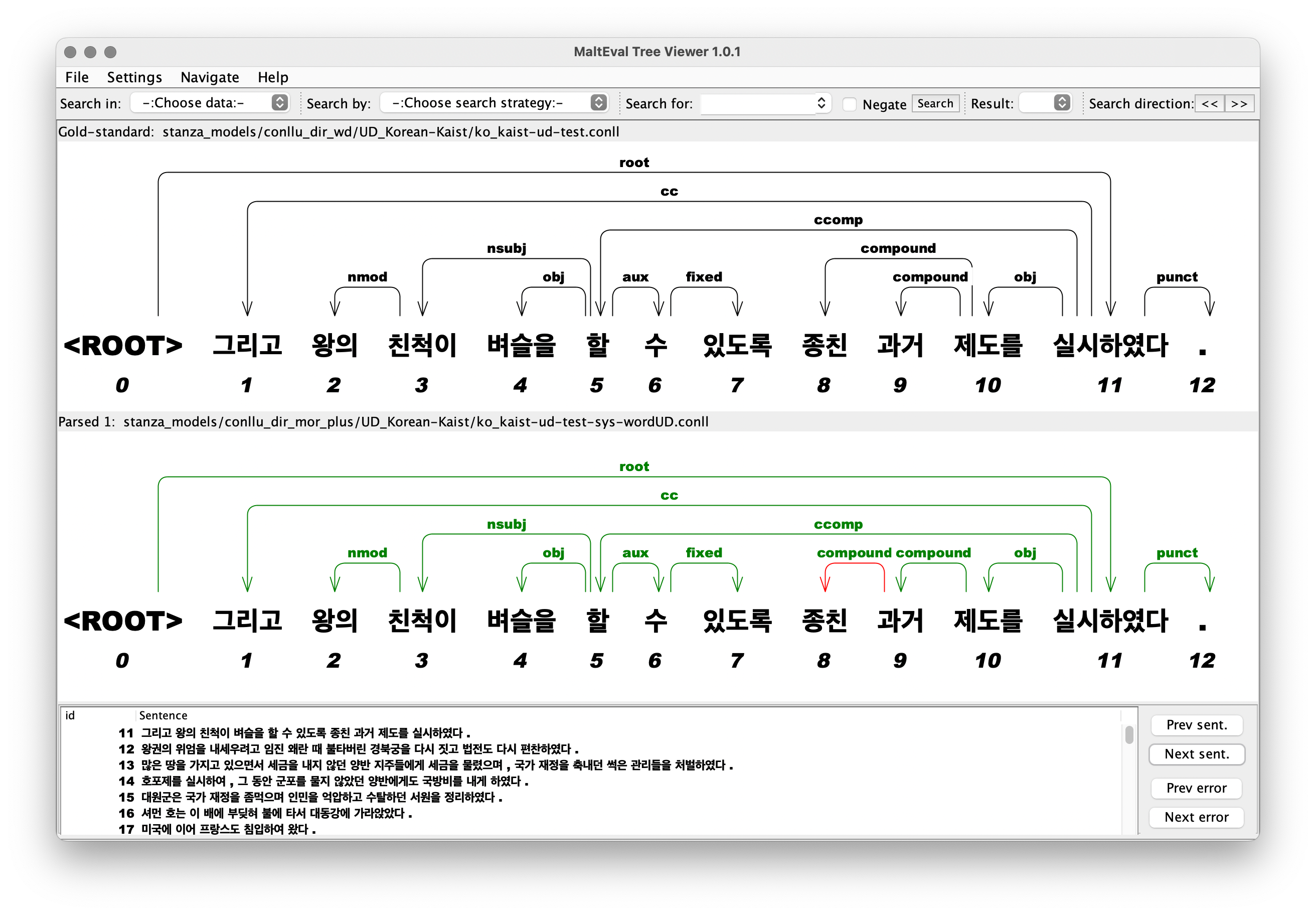} 
}
\caption{Example of the 2/1 error by MaltEval where the \texttt{gold}'s arc depth is 2 and the \texttt{system}'s depth is 1. Note that \textsc{morphUD} results are converted back to \textsc{wordUD}:  \textit{geuligo wang-ui chinjog-i byeoseul-eul hal su issdolog jongchin gwageo jedo-leul silsihayeossda} `And they introduced a clan system to make sure that the king's relatives can obtain the government position'}
\label{error-example}
\end{figure}

\begin{figure}[t]
\centering
\resizebox{.3\textwidth}{!}{
\begin{forest}
where n children=0{tier=word}{}
[NP-AJT [NP [NP [NP [실내 \\\textit{silnae}\\(`interior'),name=silnae]]
[NP [장식용 \\\textit{jangsik-yong}\\(`ornamental'),name=jangsik-yong]]]
[NP [직물 \\\textit{jikmul}\\(`textile'),name=jikmul]]]
[NP [디자이너로 \\\textit{dijaineo-ro}\\(`designer-\textsc{ajt}'),name=dijaineo-ro]]]
\draw[->] (silnae) to[out=south east,in=south] (jangsik-yong);
\draw[->] (jangsik-yong) to[out=south east,in=south] (jikmul);
\draw[->] (jikmul) to[out=south east,in=south] (dijaineo-ro);
\end{forest}
}
\caption{Compound noun with a left-skewed tree for NP modifiers in the Korean treebank} \label{np-modifiers}
\end{figure}

Figure~\ref{confusion-matrix} presents the confusion matrix for the arc depth. 
The most frequent arc depth error is 2 (\texttt{gold}) / 1 (\texttt{system}) (901 arc depth errors in \textsc{wordUD} vs. 855 in \textsc{morphUD}). 
Figure~\ref{error-example} shows an example of parsing errors generated by MaltEval \citep{nilsson-nivre:2008:LREC}. 
The parsing error shows that whereas the gold's arc requires the depth 2, the system predicts the depth 1. 
This is mainly because the analysis of compound nouns for the NP modifier in the Korean treebank prefers a left skewed tree as shown in Figure~\ref{np-modifiers} where some nouns are a verbal noun, and it plays a role as a predicate of the precedent NP modifier. 
This is a quite different from the English treebank where the right skewed tree dominates: $[_{\textsc{np}}$  $[_{\textsc{prps}}$ \textit{its}$]$ $[_{\textsc{n}}$  $[_{\textsc{nn}}$ \textit{Micronite}$]$ $[_{\textsc{n}}$  $[_{\textsc{nn}}$ \textit{cigarette}$]$ $[_{\textsc{nns}}$ \textit{filters}$]]]]]$.
This is a well-known problem when parsing the Korean treebank because it requires the semantics of the noun to distinguish between the right and the left skewed trees.
One possible remedy for this problem was to build a fully lexicalized parsing system \citep{park-EtAl:2013:IWPT}.

\section{Conclusion}

We proposed a new annotation scheme for Universal Dependencies for Korean.
We have already worked on NER, in which we outperformed the word-level representation dataset by using the morphologically enhanced dataset, and we are planning to extend our idea to the semantic role labeling task.
We are also trying to create a consortium to develop the morphologically enhanced Universal Dependencies for other morphologically rich languages such as Basque, Finnish, French, German, Hungarian, Polish, and Swedish.
All conversion scripts (\textsc{wordUD} to \textsc{morphUD}, and vice versa),  and \textsc{morphUD} datasets for \texttt{ko\_gsd} and \texttt{ko\_kaist} will be available through author’s \texttt{github} at \url{https://github.com/jungyeul/morphUD-korean}.


\bibliography{chan-etal}




\end{document}